%% file: emnlp2021.tex
\pdfoutput=1

\documentclass[11pt]{article}

\usepackage[final]{emnlp2021}

\usepackage{times}
\usepackage{latexsym}
\usepackage{todonotes}

\usepackage[T1]{fontenc}

\usepackage[utf8]{inputenc}

\usepackage{microtype}

\input{math_commands}

\usepackage{hyperref}
\usepackage{graphicx}
\usepackage{caption}
\usepackage{subcaption}
\usepackage{physics}
\usepackage[inline]{enumitem}
\usepackage{booktabs}
\usepackage{mathtools}
\usepackage{footmisc}
\usepackage{amsmath}
\usepackage{amssymb}
\usepackage{pifont}
\usepackage{xspace}

\usepackage[usestackEOL]{stackengine}

\newcommand{\citepossessive}[1]{\citeauthor{#1}'s \citeyearpar{#1}}
\def\editor{\textsc{KnowledgeEditor}\@\xspace}

\newcommand{\ie}{\emph{i.e.}\xspace}
\newcommand{\eg}{\emph{e.g.}\xspace}
\newcommand{\wrt}{wrt\xspace}

\title{Editing Factual Knowledge in Language Models}

\author{Nicola De Cao~\textsuperscript{1,2}, Wilker Aziz~\textsuperscript{1}, Ivan Titov~\textsuperscript{1,2} \\
\textsuperscript{1}University of Amsterdam,
\textsuperscript{2}University of Edinburgh \\
{\tt \{ nicola.decao, w.aziz, titov \} @uva.nl}}

\begin{document}
\maketitle

\input{0_abstract}
\input{1_introduction}
\input{2_task}

\input{3_related_work}
\input{4_method}
\input{5_experimental_setting}
\input{6_results}
\input{7_conclusions}

\bibliography{emnlp2021}
\bibliographystyle{acl_natbib}

\input{8_appendix}

\end{document}

%% file: math_commands.tex
\usepackage{amsmath,amsfonts,bm}

\def\eqref#1{equation~\ref{#1}}

\def\1{\bm{1}}

\DeclareMathAlphabet{\mathsfit}{\encodingdefault}{\sfdefault}{m}{sl}
\SetMathAlphabet{\mathsfit}{bold}{\encodingdefault}{\sfdefault}{bx}{n}

\def\gC{{\mathcal{C}}}
\def\gD{{\mathcal{D}}}

\def\gL{{\mathcal{L}}}

\def\gO{{\mathcal{O}}}
\def\gP{{\mathcal{P}}}

\def\gY{{\mathcal{Y}}}

\DeclareMathOperator*{\E}{\mathbb{E}}

\newcommand{\R}{\mathbb{R}}

%% file: 0_abstract.tex
\begin{abstract}
The factual knowledge acquired during pre-training and stored in the parameters of Language Models (LMs) can be useful in downstream tasks (\eg, question answering or textual inference). However, some facts can be incorrectly induced or become obsolete over time.  We present \editor, a method which can be used to \textit{edit} this knowledge and, thus, fix `bugs' or unexpected predictions without the need for expensive re-training or fine-tuning. Besides being computationally efficient, \editor does not require any modifications in LM pre-training (\eg, the use of meta-learning).  In our approach, we train a hyper-network with constrained optimization to modify a fact without affecting the rest of the knowledge; the trained hyper-network is then used to predict the weight update at test time.  
We show \editor's efficacy with two popular architectures and knowledge-intensive tasks: i)~a BERT model fine-tuned for fact-checking, and ii)~a sequence-to-sequence BART model for question answering. With our method, changing a prediction on the specific wording of a query tends to result in a consistent change in predictions also for its paraphrases. We show that this can be further encouraged by exploiting (\eg, automatically-generated) paraphrases during training. Interestingly, our hyper-network can be regarded as a `probe' revealing which components need to be changed to manipulate factual knowledge; our analysis shows that the updates tend to be concentrated on a small subset of components.\footnote{Source code available at \url{https://github.com/nicola-decao/KnowledgeEditor}}
\end{abstract}

%% file: 1_introduction.tex
\section{Introduction} \label{sec:intro}
\begin{figure*}
    \centering
    \begin{subfigure}[t]{0.47\textwidth}
        \centering
        \includegraphics[width=\textwidth]{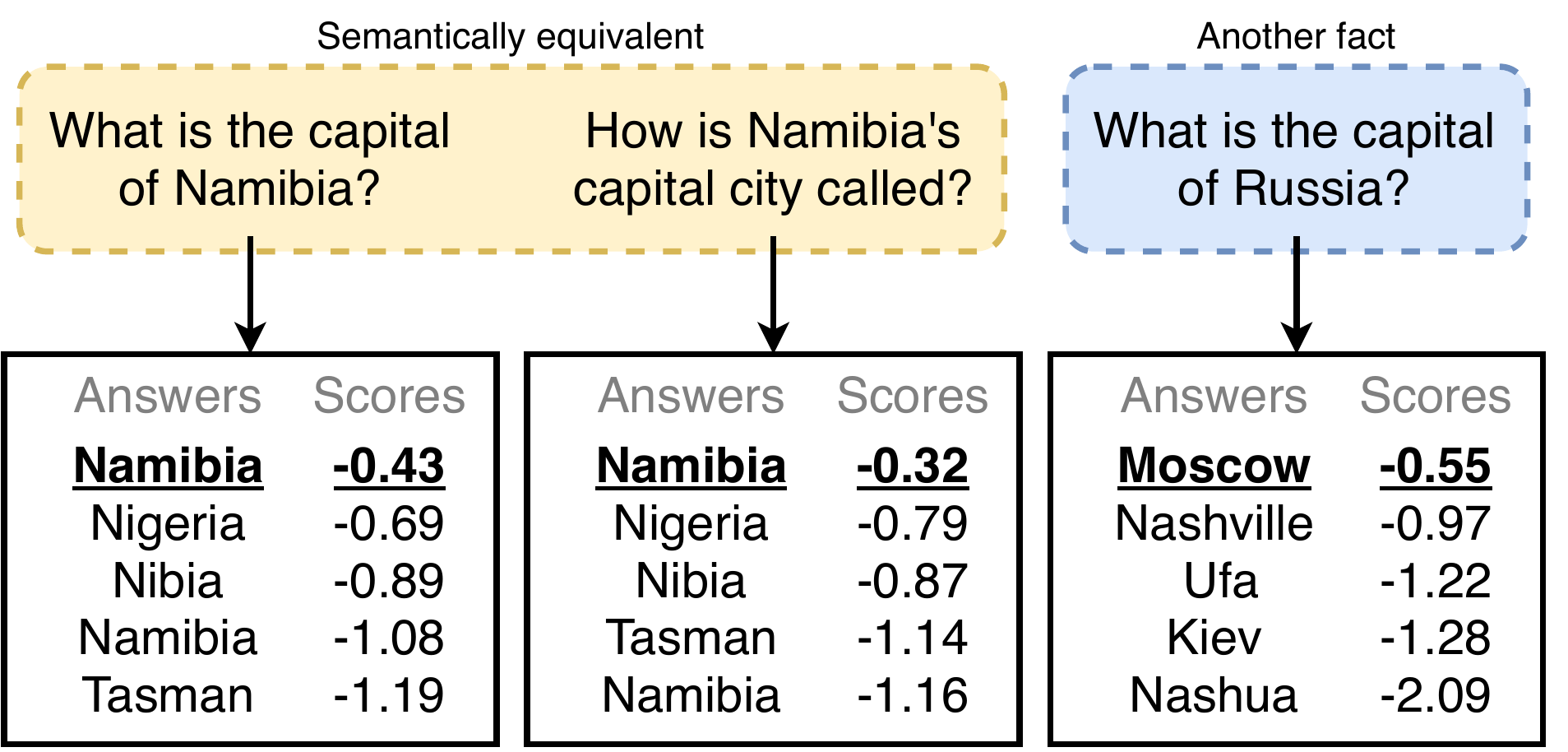}
        \caption{Model predictions before the update.}
        \label{fig:example1}
    \end{subfigure}
    \hfill
    \begin{subfigure}[t]{0.47\textwidth}
        \centering
        \includegraphics[width=\textwidth]{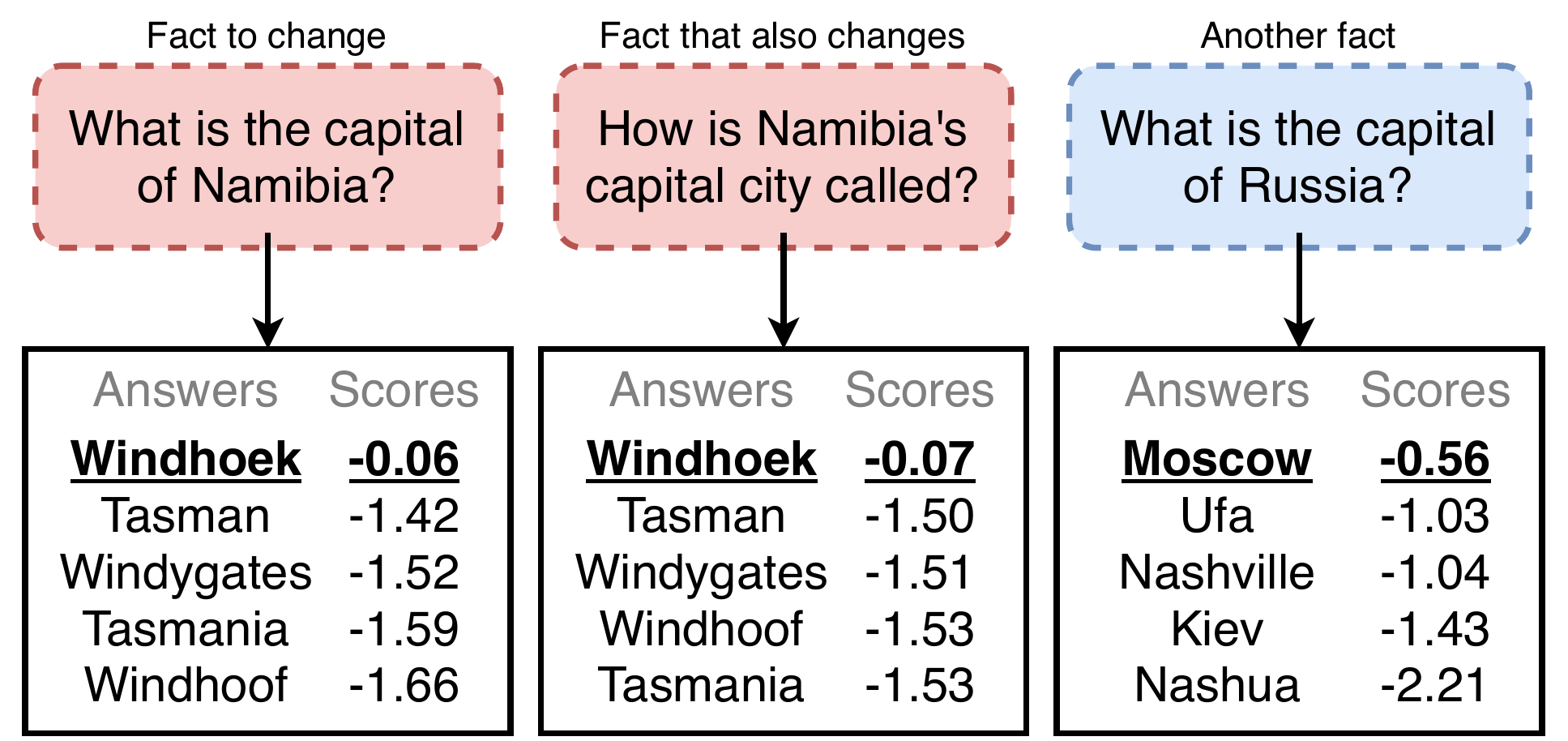}
        \caption{Model predictions with edited parameters.}
        \label{fig:example2}
    \end{subfigure}
    \caption{Predictions from a pre-trained language BART model fine-tuned for closed-book question answering. \textbf{Left:} model top-k predictions from Beam Search. \textbf{Right:} top-k after using our method conditioning on changing `\textit{What is the capital of Namibia?}' from `\textit{Namibia}' (wrong) to `\textit{Windhoek}' (correct prediction). Changing one fact also changes a semantically equivalent question and keeps the predictions from other facts the same.}
    \label{fig:example}
\end{figure*}

Using pre-trained transformer-based Language Models~\citep[LMs;][]{vaswani2017attention,devlin-etal-2019-bert,radford2019language,lewis2019bart,raffel2019exploring,brown2020language} has recently become a standard practice in NLP. Factual knowledge induced during pre-training can help in downstream tasks, but it can also be incorrect or become obsolete over time (\eg, not reflecting changes of heads of states or country populations).  Developing reliable and computationally efficient methods for bug-fixing models without the need for expensive re-training would be beneficial. %
See Figure~\ref{fig:example} for an example of revising the memory of a model that initially misremembered Namibia's capital.

\begin{figure}[t]
    \centering
    \includegraphics[width=0.48\textwidth]{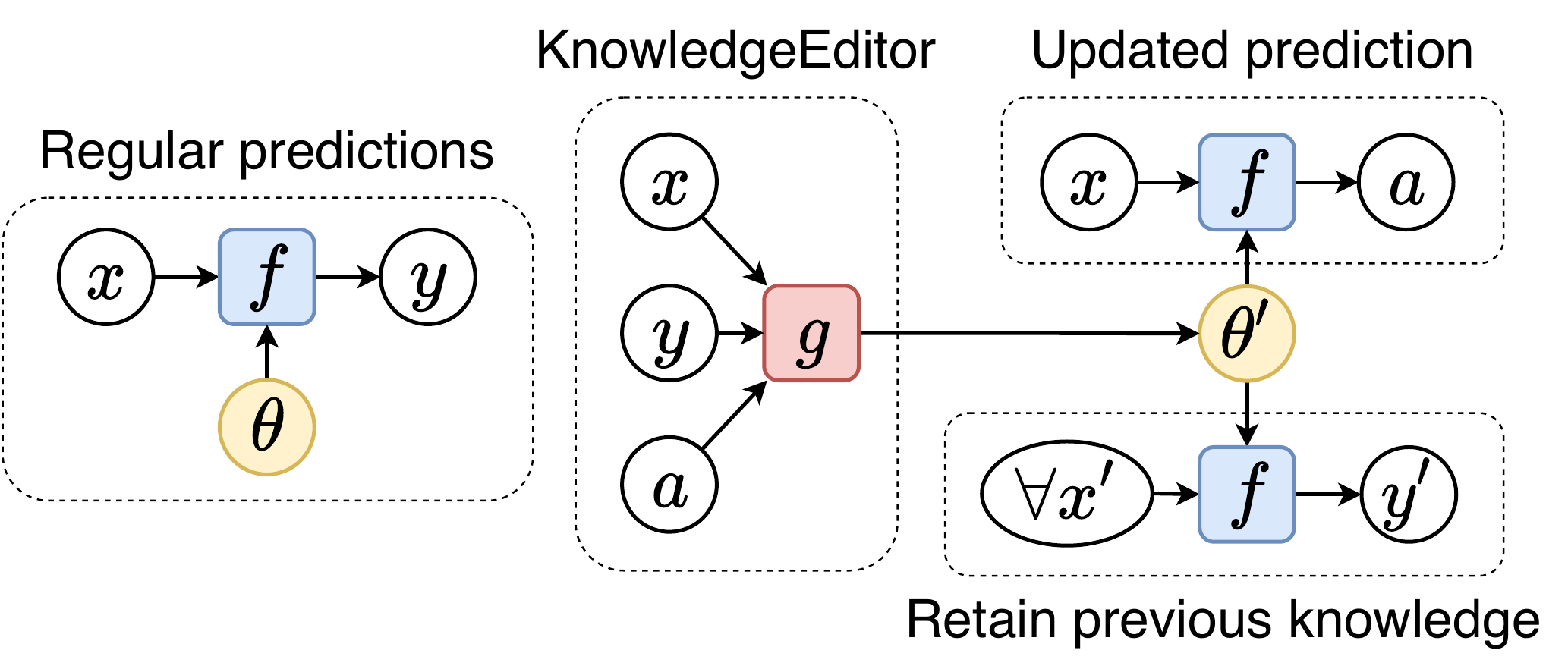}
    \caption{\textbf{Left:} a model $f$ with parameters $\theta$ prefers a prediction $y$ for input $x$ (\eg, $y$ is the mode/argmax of a discrete distribution parameterized by $f(x;\theta)$). \textbf{Right:} our method uses a hyper-network $g$ to update the parameters of $f$ to $\theta'$ such that $f(x; \theta')$ prefers an alternative prediction $a$  %
    without affecting the prediction $y'$ of \textit{any} other input $x' \neq x$. Our model \textit{edits the knowledge} about $x$ stored in the parameters of $f$.}
    \label{fig:method}
\end{figure}

Unlike conventional Knowledge Bases (KBs) that explicitly store factual knowledge, neural models implicitly memorize facts in their parameters.
One cannot easily access and interpret their computation and memories~\citep{ribeiro2016model,belinkov2019analysis,voita2019bottom,decao2020decisions}, thus, modifying their knowledge is a challenging problem.
Motivated by practical considerations,  
we formulate the following desiderata for a method aimed at tackling this problem (see Section~\ref{sec:task} for a more formal treatment):
\begin{itemize}[topsep=2pt,itemsep=-4pt]
    \item \textbf{Generality:} be able to modify a model that was not specifically trained to be editable (\ie, no need for special pre-training of LMs, such as using meta-learning);
    \item \textbf{Reliability:} be able to successfully update a specific fact without affecting the rest of the acquired knowledge;
    \item \textbf{Consistency:} the changes should be consistent across equivalent formulations of a fact (e.g., when asked to update an answer for one question, answers to its paraphrases should change accordingly).
\end{itemize}
The problem has been previously tackled in \citet{zhu2020modifying} and \citet{sinitsin2020editable}, as discussed in detail in Section~\ref{sec:related}. 
However, both  do not ensure that the edited model will be `reliable', \ie that the rest of the knowledge would not be badly affected, and that the changes are `consistent' across equivalent inputs.  Additionally,~\citepossessive{sinitsin2020editable} method requires expensive specialized training of the original network. While re-training the original network was feasible in their applications (e.g., in machine translation), it is problematic when the network is a pre-trained LM. We propose a novel method that overcomes these limitations. 

We treat editing the memories of a neural model as a \textit{learning-to-update} problem. We use an efficient parameterization of a hyper-network that is trained to update the LM parameters when provided with a single fact that needs to be modified. We do not require meta-learning, re-training or fine-tuning of the original network. We employ constrained optimization in training: we constrain the edited model to retain the same predictions as the original one regardless of the distance between the original and updated models in the parameter space. We show how this framework can be extended to incorporate (e.g., automatically-generated) paraphrases in training, further improving consistency. Figure~\ref{fig:method} shows an outline of our method.

Differently from both previous methods, we do not have to select a subset of parameters to update as we let our model learn that by itself. In fact, our hyper-network can be regarded as a ‘probe’ revealing which components of the network need to be changed to manipulate factual knowledge, \ie revealing the `causal mediation mechanisms'~\cite{vig2020causal}.
We observe that the updates end up being concentrated in a restricted set of model components, even though we do not encourage any kind of sparsity.
Interestingly, the most-updated components are different from the groups of parameters receiving large gradients (see Figure~\ref{fig:weights}).

\paragraph{Contributions} Our contributions are as follows:
\begin{itemize}[topsep=2pt,itemsep=-4pt]
    \item we define the task of knowledge editing and propose a set of evaluation metrics;
    \item we propose \editor that \textit{learns} to modify LMs memories efficiently and reliably while maintaining consistent predictions for semantically equivalent inputs;
    \item we verify that our proposed method largely meets our desiderata---while other baselines based on fine-tuning fail---testing it with different LM architectures on knowledge-intensive tasks such as fact-checking and open-domain question answering;
    \item we analyze the updates for \editor and the alternatives.
\end{itemize}
 

%% file: 2_task.tex
\section{Task}  \label{sec:task}
We want to edit the memory of a neural language model such that when, presented with an input, its output reflects a revised collection of facts. 
Unfortunately, the knowledge of a language model is typically opaque to us, being stored non-locally across a large number of parameters and architectural components. 
Thus, concretely, to operationalize the task, we seek a change in the model's parameters that affects predictions from the model only for a specific input.
For a given input $x$, the prediction $a$ made by the edited model should differ from the prediction $y$ made by the original model \emph{only} if $x$ is influenced by one of the revised facts.

\subsection{Definition}
More formally, we have a model $x \mapsto f(x;\theta)$ with trained parameters $\theta$, and a dataset of revisions $\langle x, y, a \rangle \in \gD$, \ie, $x$ is an input, $y$ is the prediction preferred by $f(x; \theta)$, and $a$ is an alternative prediction which we would like an edited version of the model to prefer.
Concretely, we keep the model architecture $f$ fixed, and seek alternative parameters $\theta'$ such that for $x$, $f(x; \theta')$ would prefer the  prediction $a$ instead of $y$ while keeping all other predictions unchanged. 
In practice, we approximate the set of `all other predictions' using a finite data set $\gO^x$ of pairs $\langle x', y' \rangle$ with $x' \neq x$.
Moreover, predictions need not be continuous nor differentiable outputs from the model; instead, they may result from an arbitrary decision rule based on $f(x; \theta)$. For example, when $f(x;\theta)$ parameterizes a discrete distribution $p_{Y|X}$ over the output space, the most standard decision rule is to output the mode of the distribution: $y = \arg\max_{c \in \gY} ~ p_{Y|X}(c|x,\theta)$.\footnote{Whereas in text classification solving this is straightforward (for $\gY$ is small), in sequence-to-sequence we resort to beam search to approximate the mode (for $\gY$ is too large or unbounded).}

\paragraph{Semantically equivalent inputs}
Optionally, for some revision $\langle x, y, a \rangle \in \gD$, we may also have a set $\gP^x$ of inputs semantically equivalent to $x$ (\eg, automatically-generated paraphrases). 
Such a set can be used in at least two ways:  i) to obtain explicit supervision for changes that should be realized in tandem with $\langle x, y, a\rangle$;
and, independently of that, ii) to evaluate whether an edited model makes consistent predictions on semantically equivalent inputs. Note that in this work we never use paraphrases at test time, only for training and evaluation of our approach; generating them at test time, while potentially helpful,  would have compromised efficiency. 

\subsection{Evaluation} \label{sec:metrics}
To test if a method $g$, producing edited parameters $\theta'$,  meets our desiderata, we measure:
\begin{enumerate}[topsep=2pt,itemsep=-4pt]
    \item \textit{success rate}: how much $g$ successfully updates the knowledge in $\theta'$, measured as accuracy of revised predictions for inputs in $\mathcal D$;
    \item \textit{retain accuracy}: how well  $\theta'$ retains %
    the original predictions of $f$, measured as accuracy \wrt input-output pairs in sets $\gO^x$; %
    \item \textit{equivalence accuracy}: how consistent the predictions of the revised model $\theta'$ are  for semantically equivalent inputs, 
    measured as accuracy of the revised predictions for all $\gP^x$;
    \item \textit{performance deterioration}: 
    how much test performance of the updated model deteriorates.\footnote{$1 -\frac{\text{accuracy of } f(\cdot; \theta')}{\text{accuracy of } f(\cdot; \theta)}$} %
\end{enumerate}
These values %
are obtained by comparing predictions of $f(\cdot; \theta)$ and $f(\cdot; \theta')$ for different subsets of inputs (\eg, $\gD$, $\gO^x$, $\gP^x$) and against different targets (\eg, gold-standard, original predictions, or alternative predictions). While these metrics are straightforward to compute in principle, some can be computationally demanding. For example, retain accuracy depends on predictions for \emph{all} inputs we have access to, which is potentially the entirety of the downstream task's validation/test data.\footnote{During training of $g$, however, we can use sub-sampling (\ie, mini batches) to approximate the metric.}  %

Previous work has evaluated similar versions of this task differently. \citet{sinitsin2020editable} measure performance deterioration and success rate but do not measure \textit{retain accuracy} nor \textit{equivalence accuracy}. 
A small performance deterioration does not guarantee high equivalence accuracy as the former is sensitive to changes in cases where the original model makes wrong decisions. 
Assessing accuracy against old or revised facts, which \citet{zhu2020modifying} also do, does not help to measure the retain accuracy.
We argue that preserving model predictions for inputs not in $\mathcal D$ is critical in production settings,  where model predictions might have been extensively analyzed and tested. 
For $x' \not\in \mathcal D$, we aim to maintain all original predictions as well as the model scores $f(x'; \theta')$ itself, effectively avoiding the need to re-calibrate the models (for example, in applications where probability estimates are used downstream).

%% file: 3_related_work.tex
\section{Related work} 
\label{sec:related}

\paragraph{Modifying transformers}
The most straightforward strategy to edit the knowledge of a model would be to re-train it on a new dataset with additional, modified, or removed facts. This is often unfeasible as LMs require large-scale expensive training that can hardly be reproduced by the most. \citet{sinitsin2020editable} propose a meta-learning approach~\citep{finn2017model} for model modification that learns parameters that are easily \textit{editable} at test time (\eg, updating the knowledge of the model requires only a few SGD steps from these learned parameters). To have a reliable method, they employ a regularized objective forcing the updated model not to deviate from the original one. This technique suffers from three main limitations: i) it requires expensive and specialized pre-training, ii) it is sensitive to many hyper-parameters (\eg, the weights of the regularizers and the subset of parameters to update), and iii) their multitask objective does not \emph{guarantee} reliability (\ie, the model is penalized for diverging from the original, rather than constrained not to).

Instead of penalizing an updated model for deviating from the original one,~\citet{zhu2020modifying} use constrained optimization. They use a less computationally expensive procedure as they \textit{re-}fine-tune on a specific downstream task (with altered data). Their method employs either an $L_2$ or $L_\infty$ constraint between the original model's parameters and the edited ones. However, a norm-based constraint on parameters ignores the highly non-linear nature of LMs and how parameters determine the outputs of the model. Indeed, a minimal change in parameter space may produce a completely different output for many datapoints leading to a potentially unreliable method. Additionally, they show the need to select a subset of parameters to be updated, which requires extra development effort. 
\citepossessive{zhu2020modifying} method is similar to Elastic Weight Consolidation~\citep{Kirkpatrick2017OvercomingCF}, a technique developed for preventing catastrophic forgetting in neural network models.

\paragraph{Knowledge in Language Models}
\citet{petroni-etal-2019-language} show that pre-trained language models recall factual knowledge without fine-tuning, which they do by feeding specific prompts to LMs. 
Hand-crafted prompts have been found not to be the best option to extract knowledge from LMs, and various solutions have been proposed to understand what LMs `know'~\citep{jiang-etal-2020-know,shin-etal-2020-autoprompt,liu2021gpt}. Additionally,~\citet{roberts-etal-2020-much} show that large models can be fine-tuned to access their internal memories to answer questions in natural language without any additional context and with surprisingly high accuracy---a setting they referred to as closed-book question answering. Although performing quite well, these models cannot reach the prediction quality of alternatives that retrieve and use context. 
Approaches that incentivize 
memorization of
factual knowledge show to be beneficial for many downstream tasks suggesting that research on methods that effectively edit the memory of a model is indeed important~\citep{zhang-etal-2019-ernie,sun2019ernie,sun2019ernie2}. Some recent hybrid approaches that use both \textit{implicit} and \textit{explicit memory} show some benefits for question answering~\citep{fevry-etal-2020-entities,verga2020facts}. Notably, language models that \textit{only} rely on internal \textit{implicit} memory are state-of-the-art for (multilingual-) Entity Linking~\citep{decao2021autoregressive,decao2021multilingual}. An effective mechanism for editing LM's implicit memory may be applicable in all these settings.

\paragraph{Causal Interventions}
Identification of minimal changes to neural networks needed to achieve a certain behaviour has been studied in the context of research in interpreting neural networks~\cite{lakretz-etal-2019-emergence,vig2020causal,elazar2021amnesic,csordas2021are}. The components which need to be updated can be interpreted as controlling or encoding the corresponding phenomena (e.g., subject-verb agreement). Much of this research focused on modifying neuron activations rather than weights and on sparse interventions (e.g., modifying one or a handful of neurons).
While far from our goals, there are interesting connections with our work. For example, our analysis of updates in Section~\ref{ss:updates}, though very limited, may shed some light on how factual knowledge is encoded in the parameters of a model.

%% file: 4_method.tex
\section{Method} \label{sec:method}

We propose to treat the task of editing the memory of a neural model as a learning problem. Instead of defining a handcrafted algorithm to compute the new parameters $\theta'$, we learn a \editor: a model that predicts $\theta'$ conditioned on an atomic fact that we want to modify. 
Concretely, \editor is a hyper-network~\citep{ha2016hypernetworks}---\ie, a neural network that predicts the parameters of another network.
Since the task requires every other prediction to stay the same---except the one we desire to change---we cast the learning task as a constrained optimization problem.

\paragraph{Optimization}
For an input $x$, changing the prediction of a model $f(\cdot; \theta)$ to $a$ corresponds to minimizing the loss $\gL(\theta; x, a)$ incurred when $a$ is the target. 
Preserving the rest of the knowledge corresponds to constraining the updated parameter $\theta'$ such that model outputs $f(\cdot; \theta')$ do not change for $x' \in \gO^x$.
Our editor $g$ is a neural network parameterized by $\phi$ which we choose by optimising the following objective for each data-point $\langle x, y, a \rangle \in \gD$:
\begin{equation} \label{eq:task}
	\begin{aligned}
	\min_\phi & \quad \sum_{\hat x \in \gP^x} \gL(\theta'; \hat x, a) \\
	\mathrm{s.t.} & \quad \gC(\theta,\theta', f;\gO^x) \le m \;,
	\end{aligned}
\end{equation}
where $\gP^{x}$ is the set of semantically equivalent inputs to $x$ (for convenience we assume it contains at least $x$), $\theta' = \theta + g(x,y,a; \phi)$, $\gC$ is a constraint on the update, and the margin $m \in \R_{>0}$ is a hyperparameter.
The constraint is used to express our desire to preserve model outputs unchanged for $x' \neq x$. Note that only $x$, but not the rest of $\gP^x$, are provided as input to the editor, as these will not be available at test time.  
In our models, $f(x; \theta)$ parameterizes a discrete distribution $p_{Y|X}$ over the output sample space $\gY$, hence we choose to constrain updates in terms 
of sums of Kullback-Leibler (KL) divergences from the updated model to the original one:
$\gC_{KL}(\theta,\theta', f;\gO^x) =$
\begin{equation}
    \sum_{x' \in \gO^x} \sum_{c \in \gY} p_{Y|X}(c|x', \theta) \log \frac{ p_{Y|X}(c|x', \theta)}{p_{Y|X}(c|x', \theta')} %
\end{equation}
The constraint pushes the updated model to predict output distributions identical to the original one for all $x' \neq x$. 
An alternative constraint we could employ is an $L_p$ norm over the parameter updates such that $g$ is optimized to make a minimal update to the original model parameter:
$\gC_{L_p}(\theta,\theta', f;\gO^x) =  \left(\sum_i |\theta_i - \theta_i'|^p \right)^{1 / p}$.
This constraint was previously used by~\citet{zhu2020modifying}.
However, such a constraint, expressed purely in parameter space and without regards to the model architecture $f$, does not directly encourage model outputs to be close to original ones in \textit{function space} (\ie, the two functions to be similar). 
Neural models are highly non-linear functions, so we do not expect this type of constraint to be effective. 
This will be empirically demonstrated in  Section~\ref{sec:results}.

\paragraph{Tractable approximations}
Non-linear constrained optimization is generally intractable, thus we employ Lagrangian relaxation~\citep{boyd2004convex} instead. 
The constraint itself poses a computational challenge, as it requires assessing KL for all datapoints in the dataset at each training step. 
For tractability, we evaluate the constraint approximately via Monte Carlo (MC) sampling (see Appendix~\ref{app:approximation} for more details).
Finally, in sequence-to-sequence models, assessing KL is intractable even for a single data point, as the sample space $\gY$ is unbounded. In such cases we approximate the computation on a subset of the sample space obtained via beam search.

\paragraph{Architecture}

Instead of predicting $\theta'$ directly, our hyper-network predicts a shift $\Delta\theta$ such that $\theta' = \theta + \Delta\theta$. 
A \textit{naive} hyper-network implementation might be over-parameterized, as it requires a quadratic number of parameters with respect to the size of the target network. Thus, we apply a trick similar to~\citet{krueger2017bayesian} to make $g$ tractably predict edits for modern large deep neural networks (\eg, BERT). Namely, $g$ makes use of the gradient information $\nabla_\theta \gL(\theta;x,a)$ as it carries rich information about how $f$ accesses the knowledge stored in $\theta$ (\ie, which parameters to update to increase the model likelihood given $a$).\footnote{A version of our hyper-network that does not use gradient information converges far too slowly.}

We first encode $\langle x, y, a \rangle$, concatenating the text with special separator and feeding it to a bidirectional-LSTM~\citep{hochreiter1997long}. Then, we feed the last LSTM hidden states to a FFNN that outputs a single vector $h$ that conditions the further computations. To predict the shift for a weight matrix $W^{n \times m} \in \theta$, we use five FFNNs conditioned on $h$ that predict vectors $\alpha, \beta \in \R^m, \gamma, \delta \in \R^n$ and a scalar $\eta \in \R$. Then 
\begin{equation} \label{eq:hypernet}
\begin{aligned}
    \Delta W = \sigma(\eta)\cdot \left( \hat\alpha \odot \nabla_W \gL(W;x,a) + \hat\beta \right) \;, \\
    \text{with} \quad \hat\alpha = \hat{\sigma}(\alpha)\;\!\gamma^\top \quad \text{and} \quad \hat\beta = \hat{\sigma}(\beta)\;\!\delta^\top \;,
\end{aligned}
\end{equation}
where $\sigma$ is the Sigmoid function (\ie, $x \mapsto (1 + \exp(-x))^{-1}$), and $\hat{\sigma}$ indicates the Softmax function (\ie, $x \mapsto \exp(x) / \sum_i \exp(x_i)$). With this formulation, the parameters for the hyper-network $\phi$ scale linearly with the size of $\theta$. An interpretation of Equation~\ref{eq:hypernet} is that an update $\Delta W$ is a gated sum of a scaled gradient of the objective %
and a bias term. The scale for the gradient and the bias are generated via an outer vector product as it allows for  efficient parameterization of a matrix with just three vectors. The gate lets the model keep some parameters unchanged. %

\paragraph{Margin annealing}
The margin $m$ is a hyperparameter and therefore fixed. However, i) it is hard to choose since it is task-dependent, and ii) it should be as small as possible. If the margin is too small, however, we risk having a small feasible set, and the model may never converge.
To address both issues, we pick some initial value for the margin and anneal it during training conditioned on validation performance: when the model successfully changes $>90\%$ of the predictions, we multiply the margin by $0.8$. We stop decreasing the margin once it reaches a desirable small value. The annealing procedure prevents the model from diverging while increasingly tightening the constraint.

%% file: 5_experimental_setting.tex
\section{Experimental Setting} \label{sec:setting}

We aim to evaluate the effectiveness of \editor comparing to baselines on knowledge-intensive tasks where the importance of modifying the memory of a large LM has a broad impact. We then test our method on closed-book fact-checking and closed-book question answering with the metrics proposed in Section~\ref{sec:metrics}.

\subsection{Baselines}
We compare against two baselines: i) fine-tuning and ii) the method proposed by~\citet{zhu2020modifying}. Fine-tuning corresponds to using standard gradient descent, minimizing the loss for the fact/prediction we want to revise. 
For this, we follow~\citet{sinitsin2020editable} and employ RMSProp~\citep{tieleman2012lecture}.\footnote{We tried alternatives, RMSProp was the most effective.} %
We set the learning rate to $10^{-5}$ and stop upon successfully changing the output of the model or having reached a maximum of $100$ gradient steps.
\citepossessive{zhu2020modifying} method extends fine-tuning with an $L_\infty$ constraint on parameters.\footnote{
We search the hyper-parameter for the penalty $m\in \{10^{-3},5\times 10^{-4},10^{-4}, 5\times 10^{-5},10^{-5}\}$ selecting the best based on the sum of success rate and retain accuracy.}
Following both~\citet{sinitsin2020editable} and~\citet{zhu2020modifying} we report these baselines
fine-tuning all parameters or just a subset of them. We limit the search to selecting entire layers and base our decision on performance on a subset of the validation set.
Note that selecting a subset of parameters for update requires an extensive search, which \editor dispenses with by automatically learning it.

\input{main-table}

\subsection{Models and data}
We evaluate on closed-book fact-checking (FC) fine-tune a BERT base model~\citep{devlin-etal-2019-bert} on the binary FEVER dataset~\citep{thorne-etal-2018-fever} from KILT~\citep{petroni2020kilt}. 
We also evaluate on a task with a more complex output space: closed-book question answering (QA). For that we fine-tune a BART base model~\citep{lewis2019bart} with a standard seq2seq objective on the Zero-Shot Relation Extraction (zsRE) dataset by~\citet{levy-etal-2017-zero}. We evaluate on this dataset because it is annotated with human-generated question paraphrases that we can use to measure our model's robustness to semantically equivalent inputs.
We create alternative predictions for FC simply flipping the labels, whereas for QA we pick all hypotheses enumerated via beam search except the top-1. The latter ensures high-probability outcomes under the model distribution. We generate semantically equivalent inputs with back-translation. See Appendix~\ref{app:setting} for technical details on models and data collection.

%% file: main-table.tex
\begin{table*}[t]
	\centering
	\small
	\begin{tabular}{l cccc|cccc}
		\toprule
		& \multicolumn{4}{c}{\textbf{Fact-Checking}} &  \multicolumn{4}{c}{\textbf{Question Answering}}  \\
		\cmidrule(r{2.5pt}){2-5} \cmidrule(l{2.5pt}){6-9}
		\textbf{Method} & \Longstack{\textbf{Success} \\ \textbf{rate $\uparrow$}} & \Longstack{\textbf{Retain} \\ \textbf{acc $\uparrow$}} & \Longstack{\textbf{Equiv.} \\ \textbf{acc $\uparrow$}} & \Longstack{ \textbf{Perform.} \\ \textbf{det $\downarrow$}} & \Longstack{\textbf{Success} \\ \textbf{rate $\uparrow$}} & \Longstack{\textbf{Retain} \\ \textbf{acc $\uparrow$}} & \Longstack{\textbf{Equiv.} \\ \textbf{acc $\uparrow$*}} & \Longstack{\textbf{Perform.} \\ \textbf{det $\downarrow$}} \\
		\midrule
		Fine-tune {\small (1st layer)} &      100.0 &  99.44 &  42.24 &  0.00  &  98.68 & 91.43 & 89.86 / 93.59 & 0.41 \\
		Fine-tune {\small (all layers)} &      100.0 &  86.95 &  95.58 &  2.25 & 100.0 & 67.55 & 97.77 / 98.84 & 4.50 \\
		\citeauthor{zhu2020modifying} {\small (1st layer)} &      100.0 &      99.44 &            40.30 &           0.00 & 81.44 & 92.86 & 72.63 / 78.21 & 0.32 \\
		\citeauthor{zhu2020modifying} {\small (all layers)} &      100.0 &       94.07 &            83.30 &            0.10 & 80.65 & 95.56 & 76.41 / 79.38 & 0.35\\
		\midrule
		Ours $\gC_{L_2}$ &       99.10 &  45.10 &  99.01 &  35.29 & 99.10 & 46.66 & 97.16 / 99.24 & 9.22\\
		\midrule
		\editor  &       98.80 &  98.14 &  82.69 &  0.10 & 94.65 & 98.73  & 86.50 / 92.06 &  0.11 \\
		~ + loop$^\dagger$ &   100.0 &  97.78 &  81.57 &  0.59 & 99.23 & 97.79 & 89.51 / 96.81 & 0.50 \\
		~ + $\gP^x$ $^\ddagger$ &       98.50 &  98.55 &  95.25 &  0.24 & 94.12 & 98.56 & 91.20 / 94.53 & 0.17 \\
		~ + $\gP^x$ + loop$^\ddagger$ & 100.0 &  98.46 &  94.65 &  0.47 & 99.55 & 97.68 & 93.46 / 97.10 & 0.95\\
		\bottomrule
	\end{tabular}
	\caption{Accuracy scores on fact-checking and question answering for the metrics presented in Section~\ref{sec:metrics}. *We report both the accuracy on the set of generated paraphrases (left) and human-annotated (right).$^\dagger$Apply updates in a loop, stopping when the update is a success or when reaching a maximum number of iterations (only at test time). $^\ddagger$Using paraphrases (semantically equivalent inputs) as additional supervision (only at training time).}
	\label{tab:fever-zsre}
\end{table*}

%% file: 6_results.tex
\section{Results} \label{sec:results}
Table~\ref{tab:fever-zsre} reports the main results for fact-checking and question answering. 
Overall, \editor achieves high performance in all metrics. Some other methods also achieve high accuracy in some metrics but always sacrificing others (\ie, never meeting all our desiderata at once). 

We compare methods along different metrics (as opposed to a single one), as there is no way to precisely determine the importance of each of these metrics. To gather more insight, we compute their stochastic convex combination with coefficients sampled from a Dirichlet distribution (with $\alpha=1$ to ensure a very diverse set of combinations) and report in Figure~\ref{fig:heatmap} in Appendix~\ref{app:additional} an estimate of the probability that a system outperforms another across $1,000$ such combinations. 
The probability of our full method to outperform all baselines is very high for both FC and QA ($\approx\!97\%$ and $\approx\!88\%$, respectively). In Figure~\ref{fig:dirichlet} in Appendix~\ref{app:additional}, we show the distributions of the combined scores (\ie, the raw data for the approximation reported in Figure~\ref{fig:heatmap}). We then analyze different aspects of our method and the baselines. %

\subsection{Success rate}
Every method achieves an almost perfect success rate on fact-checking. All methods but ours apply updates in a loop, stopping either when the new model is successfully updated or after reaching a maximum number of iterations. The success rate for \editor is not $100\%$ because we do not apply more than one update even in case of failure. To this end, we also show an experiment with our method with multiple updates within a \textit{loop} employing the same stopping criteria as the baselines. Note that we apply this only at test time (\ie, we do not train for multiple updates). When applying multiple updates also our method reaches a $100\%$ success rate on fact-checking and almost perfect accuracy ($>99\%$) for QA.\footnote{Even if we do not train for multiple subsequent updates, its success opens the possibility to add this at training time. 
We leave the exploration of this technique to future work.}

Closed-book QA is a more challenging task since the output space is text and not just a binary label. In this setting, \editor achieves high accuracy ($\approx\!95\%$ or $>99\%$ with the \textit{loop}). 
Among all methods, \editor gets the best success rate while also obtaining the best retain accuracy. 
In QA,~\citepossessive{zhu2020modifying} method does not reach a good success rate ($\approx\!80\%$). We searched hyperparameters for their method also to have high retain accuracy, and indeed that is higher than regular fine-tuning. 
However, unlike fact-checking, regular fine-tuning for QA gets almost perfect scores but at the expense of the retain accuracy. Sequence-to-sequence models are more sensitive to a slight parameter shift. This happens because minor changes may completely alter the top-k prediction from beam search (in the case of QA). Differently, in a binary classifier (in the case of FC) the probability of a prediction can change substantially without crossing the decision boundary (usually set at 0.5 when not calibrated).

\begin{figure*}[t]
    \centering
    \begin{subfigure}[b]{0.32\textwidth}
        \centering
        \includegraphics[width=\textwidth]{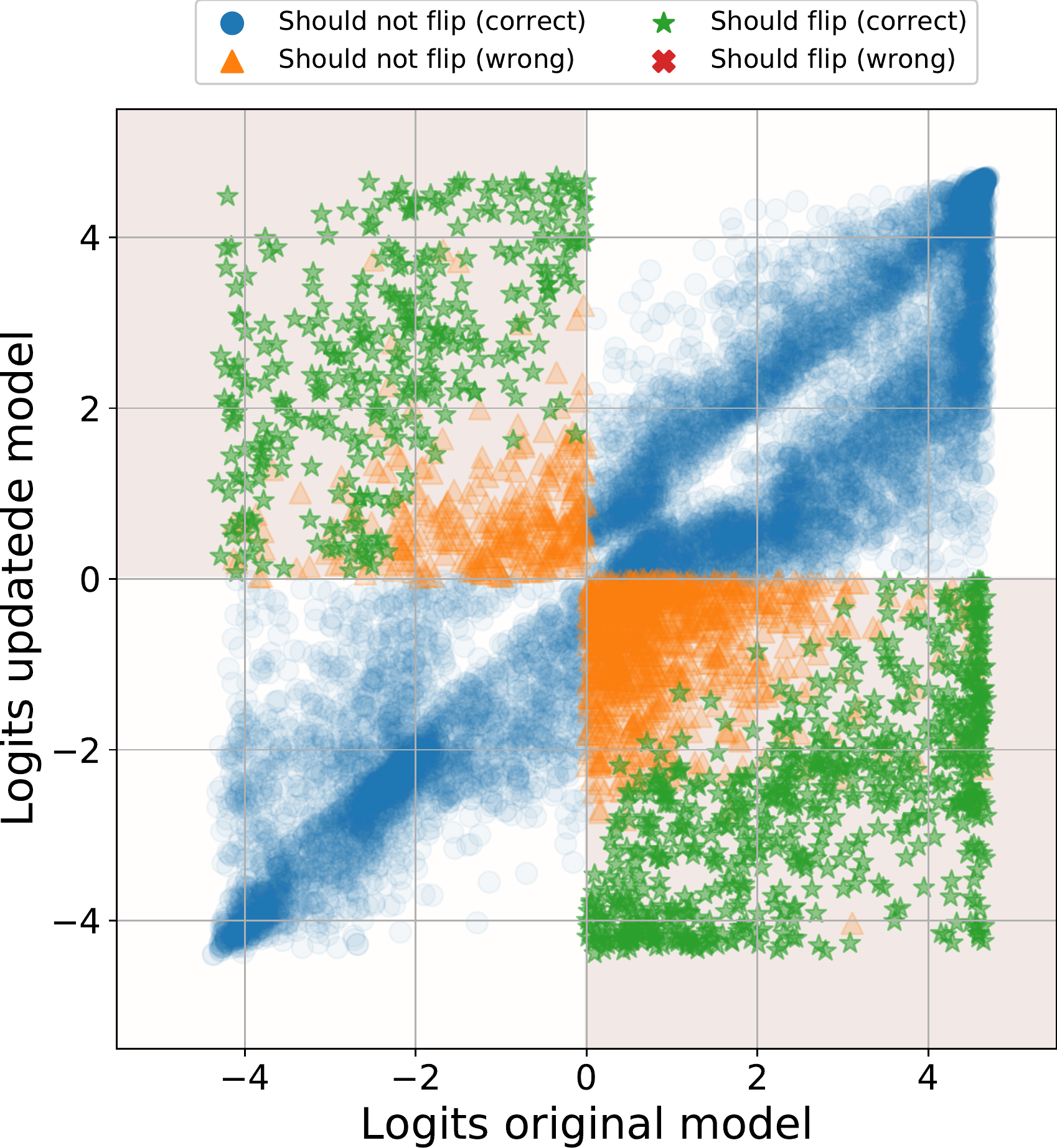}
        \caption{Fine-tune (all layers).}
        \label{fig:fever_logits1}
    \end{subfigure}
    \hfill
    \begin{subfigure}[b]{0.32\textwidth}
        \centering
        \includegraphics[width=\textwidth]{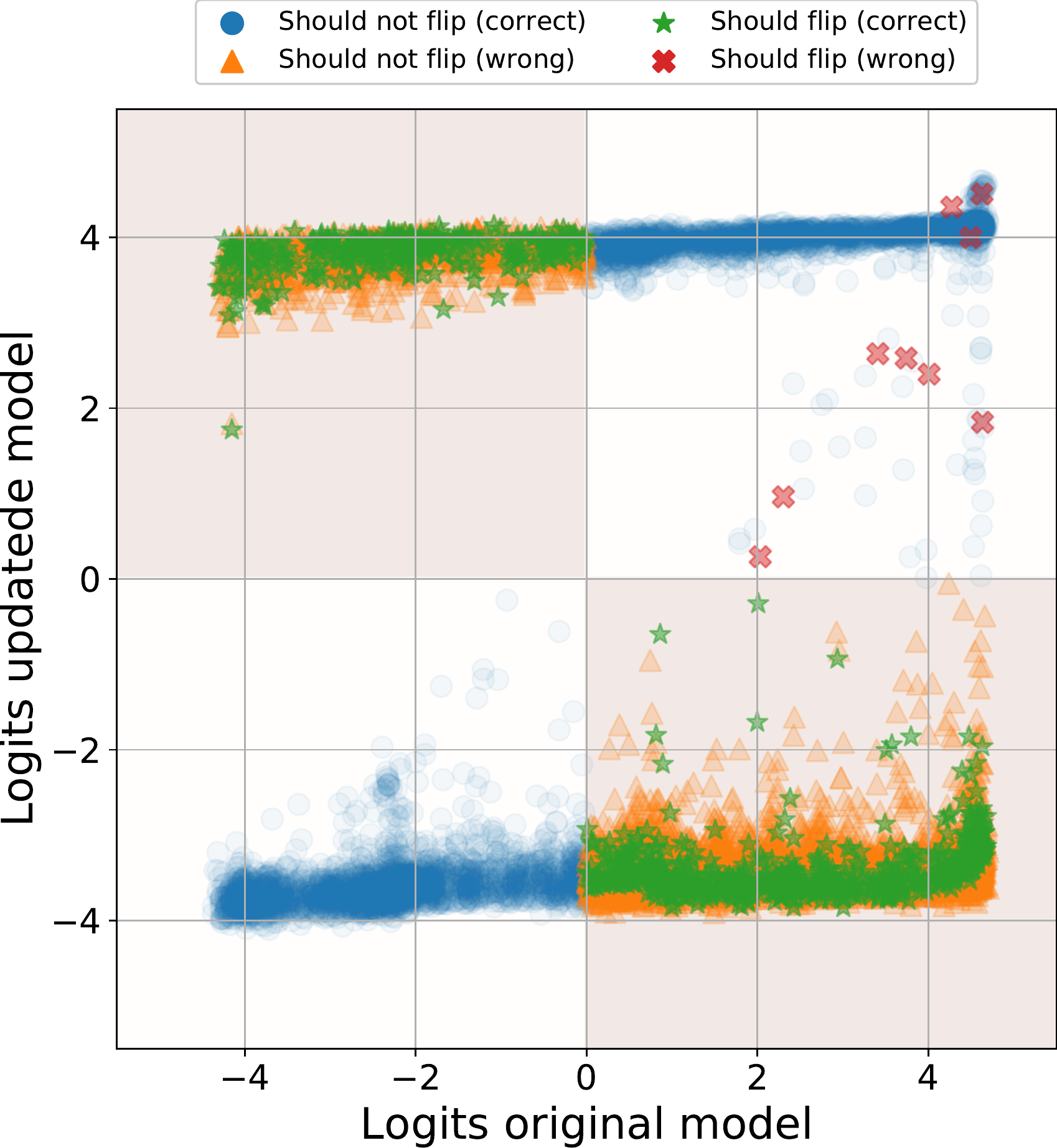}
        \caption{$\gC_{L_2}$.}
        \label{fig:fever_logits2}
    \end{subfigure}
    \hfill
    \begin{subfigure}[b]{0.32\textwidth}
        \centering
        \includegraphics[width=\textwidth]{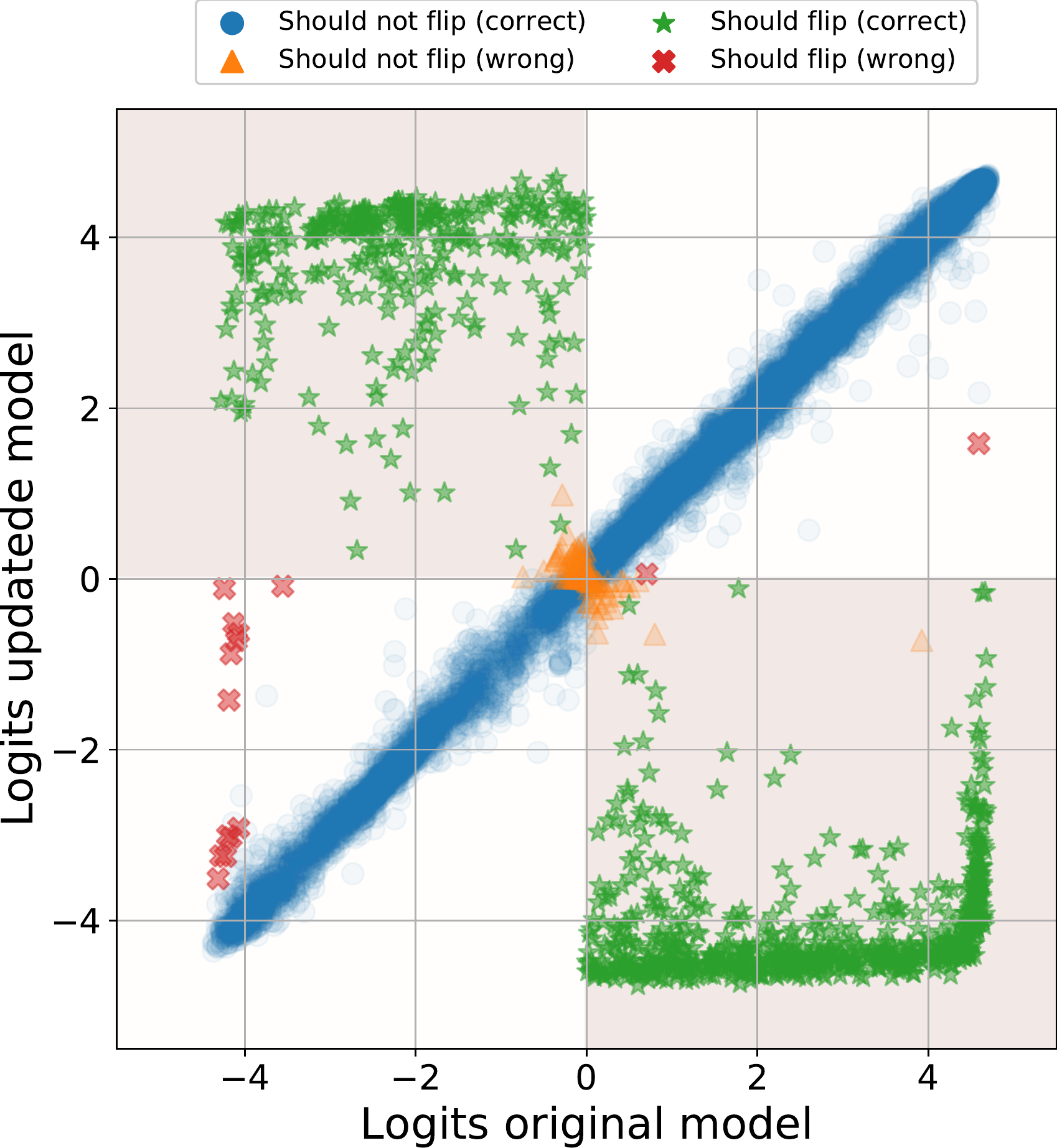}
        \caption{Ours $\gC_{KL}$ with $\gP^x$.}
        \label{fig:fever_logits3}
    \end{subfigure}
    \caption{Distribution of logits of the original model and updated model on FEVER. Fine-tuning all layers (a) leads to many errors, and the probability of the predictions does not stay the same even when they do not cross the decision boundary. $\gC_{L_2}$ (b) successfully flips labels, but it does not force the predictions to stay the same. For our full method, $\gC_{KL}$ with $\gP^x$ (c), errors are mainly concentrated around the origin where the model is uncertain, and small perturbations make logits to cross the decision boundary. \textit{Better view with colors.}}
    \label{fig:fever_logits}
\end{figure*}

\subsection{Retaining previous knowledge}
\editor maintains the predictions in the validation set almost perfectly (retain accuracy is $\approx\!98\%$ for both FC and QA).
Conversely, as expected, our method with $\gC_{L_2}$ has very low retain accuracy (always $<50\%$). $\gC_{L_2}$ suffers from catastrophic forgetting because it does not enforce the updated model to be close to the original one in function space (\ie, the two functions to be similar) but just in parameter space.

Fine-tuning all layers is successful but it affects the previously acquired knowledge negatively: retain accuracy is $\approx\!87\%$ and $\approx\!68\%$ for FC and QA, respectively, while performance deterioration in $\approx\!2\%$ and $\approx\!4\%$.
Fine-tuning a single layer is more effective as it prevents over-fitting (the best model updates the 1st layer in both FC and QA). However, in FC the updated model does not generalize on semantic equivalent inputs: the accuracy on paraphrases is much lower even than versions of our methods which do not use paraphrases in training ($42\%$ vs. $>81\%$), and even more so when compared to those which use them ($>94\%$). 

Fine-tuning with~\citepossessive{zhu2020modifying} method does not affect performance for FC much, which is not surprising since standard fine-tuning already gets almost perfect scores. Differently, in the QA setting, using their constrained optimization boosts the retain accuracy (up to $+4\%$ to normal fine-tuning) but at the cost of a low success rate ($\approx\!80\%$ where fine-tuning gets the perfect score).

\subsection{Accuracy on paraphrases}
We evaluate our method both with and without the additional supervision of paraphrases to improve generalization---that corresponds to have $\gP^x$ as the set of paraphrases of $x$ or $\gP^x=\{x\}$ in Equation~\ref{eq:task}, respectively. Without this additional supervision, \editor is already competitive in equivalence accuracy. However, employing this additional supervision is clearly beneficial on both tasks: we get the same success rate and re-train accuracy but equivalence accuracy improves by $>70\%$ on FC and $>30\%$ on QA, respectively (for generated paraphrases).
In FC, although fine-tuning of a single layer proved to be optimal in terms of success rate and retain accuracy, it performs poorly for paraphrases. That is the model successfully updates the prediction of a particular datapoint, but does not update predictions of paraphrases. This indicates that fine-tuning to edit the knowledge of a model does not generalize well, and it \textit{overfits} to specific inputs. On QA, also~\citet{zhu2020modifying} performs poorly compared to our or other methods.

When other methods perform on par or better than ours on paraphrases, they do not have good retain accuracy (\eg, see QA fine-tuning on Table~\ref{tab:fever-zsre}). Fine-tuning on QA seems to generalize better than on FC, but does not preserve previous knowledge.
In Table~\ref{tab:fever-zsre} we also report both the accuracy on the set of generated and human-generated paraphrases. 
Surprisingly, the
scores on human-generated paraphrases are higher. We speculate that this happens because automatic paraphrases are sometimes not semantically equivalent or fluent.  

\begin{figure*}[t]
    \centering
    \begin{subfigure}[b]{0.3\textwidth}
        \centering
        \includegraphics[width=\textwidth]{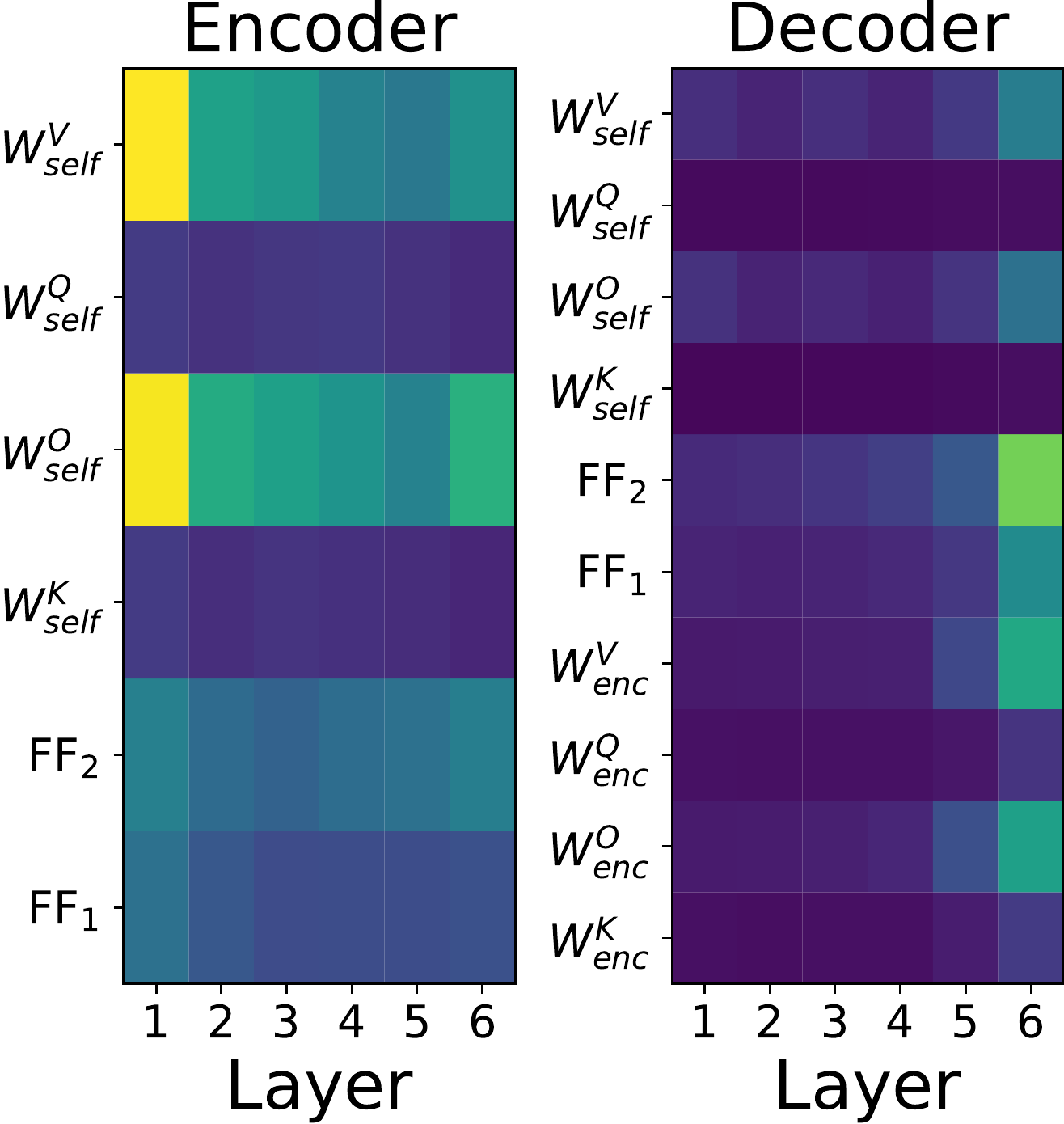}
        \caption{Gradients.}
        \label{fig:weights1}
    \end{subfigure}
    \hfill
    \begin{subfigure}[b]{0.3\textwidth}
        \centering
        \includegraphics[width=\textwidth]{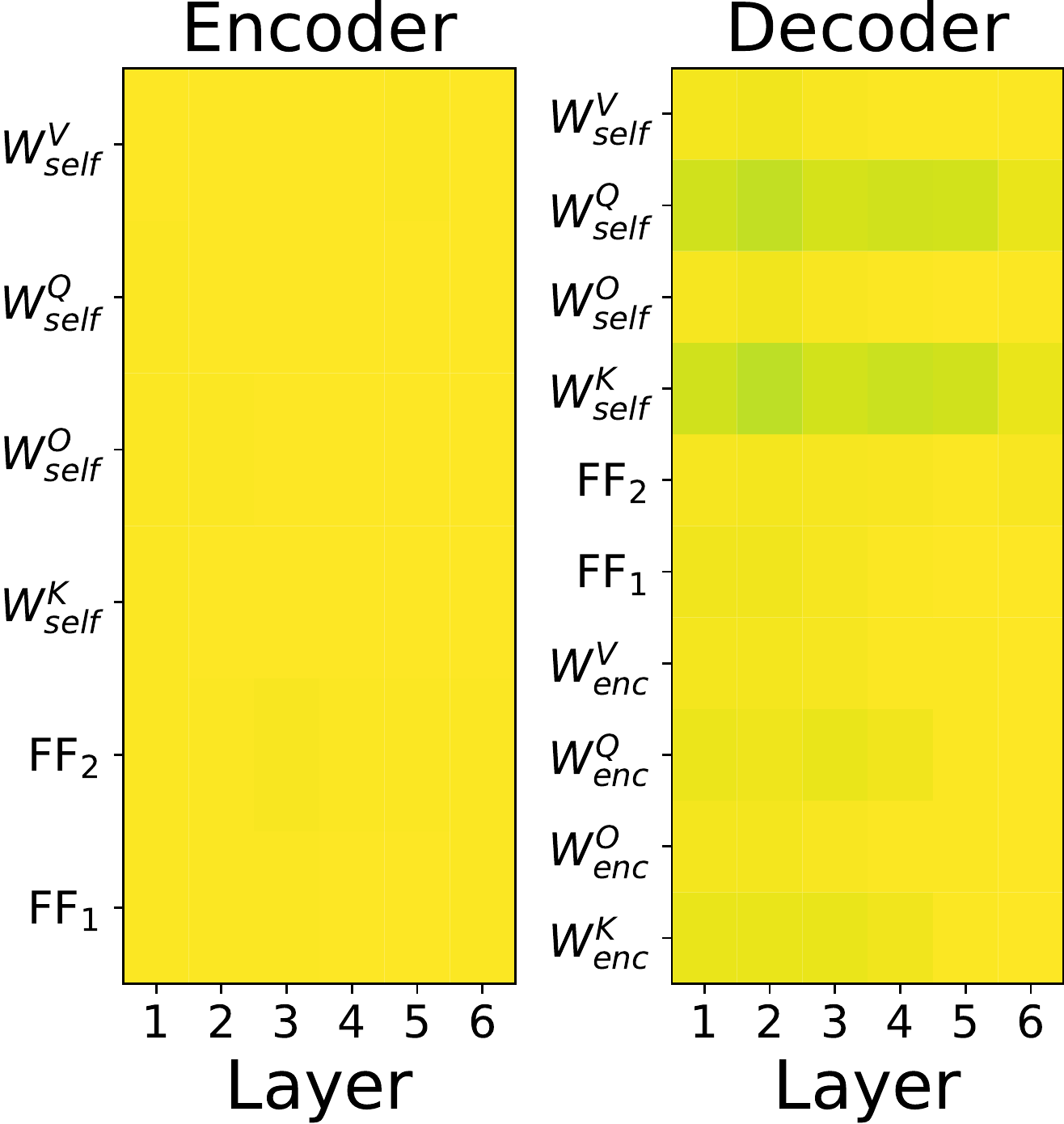}
        \caption{Fine-tune (all layers).}
        \label{fig:weights2}
    \end{subfigure}
    \hfill
    \begin{subfigure}[b]{0.3\textwidth}
        \centering
        \includegraphics[width=\textwidth]{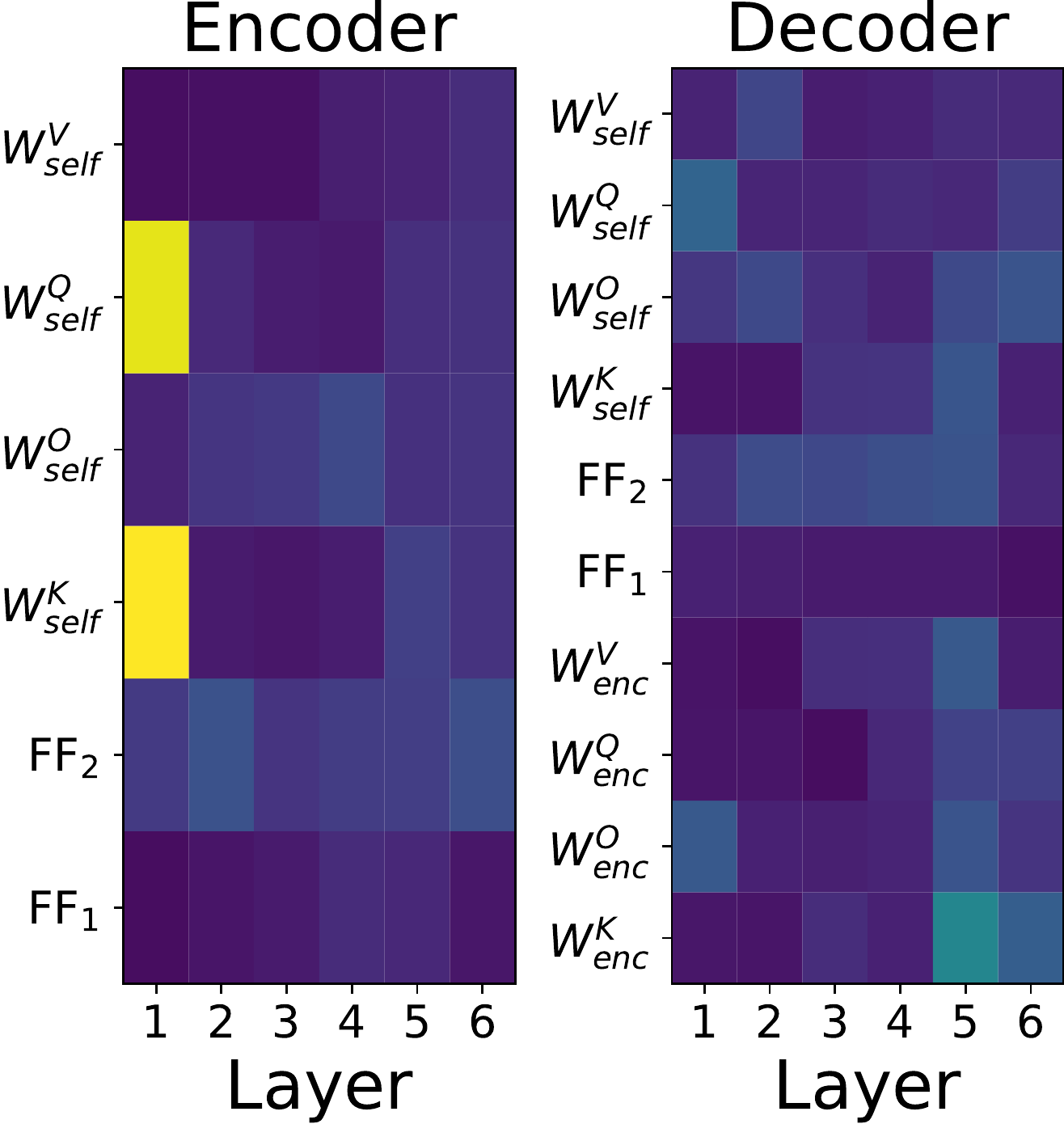}
        \caption{\editor + $\gP^x$.}
        \label{fig:weights3}
    \end{subfigure}
    \caption{Average normalized magnitude of updates on weight matrices across layers for the QA experiment. Fine-tuning updates all layers uniformly while our updates are more sparse.}
    \label{fig:weights}
\end{figure*}

\subsection{Analysis of model updates}
\label{ss:updates}
In Figure~\ref{fig:fever_logits} we plot the distribution of logits of the original and updated model on FC for different methods. With an ideal method, all logits before and after an update have to stay the same (except the ones we want to change). From that figure, we can see distributions of different types of errors such as datapoints whose predictions were mistakenly flipped (from true to false or the other way around). These errors are mostly concentrated around the origin, where small perturbations make logits cross the decision boundary. 
When fine-tuning all layers, we can see a clear impact on logits, they undergo a lot of change (\ie, points do not concentrate around the diagonal). Indeed, fine-tuning makes many datapoints cross the decision boundary and their probabilities to change from the original ones. The failure of $\gC_{L_2}$ is visible in Figure~\ref{fig:fever_logits2} as this method preserves almost none of the previous predictions. Instead \editor preserves almost all of the predicted labels as well as their probabilities (most datapoints in Figure~\ref{fig:fever_logits3} stay on the diagonal).

We also report visualizations of the average weight updates for the QA experiment in Figure~\ref{fig:weights}. We report the setting with additional supervision from paraphrases (but the heatmaps are similar without them). %
There are three main observations from this plot. First, gradients are mostly concentrated on the first encoder layer and the last decoder layer. Gradients explain why the best subset of parameters to update is the first layer. Secondly, fine-tuning does not preserve gradient magnitudes and updates the whole model almost uniformly. That happens because of the optimizer's adaptive learning rate that initially erases the gradient direction. The gradient direction plays a role only after a couple of gradient steps, but most of the time, the method only needs one step to modify its knowledge. Lastly, our updates are sparser and are not consistent with the gradient for changing the predictions. That indicates that our method learns to use the gradient in a meaningful way (\ie ignoring some directions or manipulating its magnitude). It is surprising that the knowledge manipulation seems to be achieved by primarily modifying parameters affecting the shape of the attention distribution ($W_{self}^K$ and $W_{self}^Q$) rather than, \eg, values ($W^V_{self}$). As we discussed, the hyper-network may be regarded as a probe providing insights about the mechanism used by the model to encode the knowledge~\cite{vig2020causal}. For example, the focus on the bottom layer is already intriguing, as it contrasts with claims that memorization happens in top layers of image classification models~\cite{stephenson2021geometry}, hinting at substantial differences in the  underlying memorization mechanisms in NLP and vision.  Proper investigation is however outside of the scope of this study. See Appendix~\ref{app:additional} for some additional analysis.

%% file: 7_conclusions.tex
\section{Conclusions} \label{sec:conclusions}

In this work, we explore the task of editing the factual knowledge implicitly stored in the parameters of Language Models. For this task, we formally define desiderata, the objective, and a set of metrics to measure the efficacy of different methods. We concretely evaluate that on two benchmarks based on closed-book fact-checking and question answering.
We propose \editor, a method based on a hyper-network that \textit{learns} to modify implicit knowledge stored within LM parameters efficiently and reliably. We provide comprehensive evaluations for our models against different variants of fine-tuning demonstrating the advantage of our approach. The magnitude of the updates predicted by our method may unfold the mechanisms used by the LMs to encode factual knowledge; we leave such investigation for future work.

\section*{Ethical Considerations}
Technology built upon pre-trained LMs inherits some or all of their potential harms~\citep{bender2021dangers}.
Our technology for editing the knowledge of LMs does not exacerbate their potential harms and can, in fact, be used to mitigate harms, as models can be corrected once problems are discovered. However, we note that malicious uses of our knowledge editor are possible. For example, malicious agents may use the techniques presented in this work to inject incorrect knowledge into LMs.  

\section*{Acknowledgments}
The authors want to thank Michael Schlichtkrull, Lena Voita and Luisa Quarta for helpful discussions and support. This project is supported by SAP Innovation Center Network, ERC Starting Grant BroadSem (678254), the Dutch Organization for Scientific Research (NWO) VIDI 639.022.518, and the European Union's Horizon 2020 research and innovation programme under grant agreement No 825299 (Gourmet).

%% file: 8_appendix.tex
\clearpage
\appendix

\section{Relaxation and Approximation of Constrained Optimization} \label{app:approximation}

Given a objective to minimize in the form of
\begin{equation}
	\begin{aligned}
		\min_\phi & \quad \E_{x \sim p(x)} \left[ f(x,\theta) \right] \\
		\mathrm{s.t.} & \quad \frac{1}{|\gY|}\sum_{x \in \gY} \gC(y,\theta) \le m \;,
	\end{aligned}
\end{equation}
can be solved with Lagrangian relaxation~\citep{boyd2004convex} using a multiplier $\alpha \in \R_{\geq 0}$
and be approximated by sampling $y \sim p(y)$ to
\begin{equation} \label{eq:approximation}
	\min_\phi\max_\alpha \; f(x,\theta) + \alpha \cdot  \left(\gC(y,\theta) - m \right) \;.
\end{equation}
Equation~\ref{eq:approximation} can be evaluated with automatic differentiation and optimized via gradient descent.

\section{Experimental setting} \label{app:setting}

\subsection{Fact-checking}
We evaluate on closed-book fact-checking (FC) using the binary FEVER dataset~\citep{thorne-etal-2018-fever} from KILT~\citep{petroni2020kilt}. 
FEVER has 104,966 training and 10,444 validation instances respectively.
For every input claim $x$, the model predicts the probability $f(x;\theta)$ that it may be true. This is done \textit{without} retrieving any evidence from a corpus, instead, just by relying on the knowledge accumulated during pre-training and encoded in its own parameters---this is similar to~\citet{lee-etal-2020-language} that investigate closed-book and zero-shot FC using masked-LMs. Concretely, we ask the LM to perform binary classification. We fine-tune a BERT base model~\citep{devlin-etal-2019-bert} with an additional linear layer on top that maps the hidden state corresponding to the BOS (beginning of a sentence) token to the probability of the positive label. Given the available supervision, we train the architecture to maximize the model likelihood penalized by entropy regularization and weight decay. The final model has an accuracy of $77.1\%$.\footnote{This is comparable with what reported by~\citet{petroni2020kilt} for a larger BART model.}

\subsection{Question answering}
We also evaluate on a task with a more complex sample space: closed-book question answering (QA). Here QA is treated as a sequence-to-sequence problem from question to answer without retrieving nor providing any evidence~\cite{roberts-etal-2020-much}.
This, as in FC, emphasises the role of the knowledge acquired in pre-training and encoded in the parameters of the model.
For this task, we used the Zero-Shot Relation Extraction (zsRE) dataset by~\citet{levy-etal-2017-zero}. We prefer zsRE to other popular QA datasets such as SQuAD~\citep{rajpurkar-etal-2016-squad}, Natural Questions~\citep{kwiatkowski-etal-2019-natural} or TriviaQA~\citep{joshi-etal-2017-triviaqa} because it is annotated with human-generated question paraphrases that we can use to evaluate our model's robustness to semantically equivalent inputs. zsRE is specifically constructed not to have relation overlaps between training and test (\ie it is zero-shot). 
We re-split the dataset to have the same distribution in training and test splits---we are not interested in zero-shot specifically, so we avoid the additional complexity it entails.
The original zsRE dataset has 147,909 training and 3,724 validation instances respectively. After re-splitting and employing all paraphrases, we have  244,173 training and 27,644 validation instances respectively.
For this task, we fine-tune a BART base model~\citep{lewis2019bart} with a standard seq2seq objective, \ie, maximizing the model likelihood given the observed output sequence~\citep{sutskever2011generating,sutskever2014sequence} and regularized with dropout~\citep{JMLR:v15:srivastava14a} and label smoothing~\citep{szegedy2016rethinking}. The final model has an accuracy (exact match between model prediction and gold standard) of $22.1\%$.\footnote{This is more than reported by~\citet{petroni2020kilt} on the original split of zsRE. That is because the original split aims at zero-shot evaluation, while we have an overlap of relation types between training and validation sets.}

\subsection{Generating alternative predictions}
Generation of alternative predictions is task-dependent as it requires producing a plausible substitute target for a given input---\eg, if we need to edit the knowledge about a head of a state, a plausible substitute label should be a person, not a random (even if well-formed) string. Fact-Checking is straightforward: we simply flip the label, as it is binary classification. For QA, we exploit high-probability outcomes under the model distribution as a proxy to plausible revisions. In particular, we pick all hypotheses enumerated via beam search except the top-1.\footnote{This does not always guarantee that the alternative predictions have the same semantic type as the original one, but it is likely since the model assigns high probability to them.}

\subsection{Semantically equivalent inputs}
We would like the updated model to be consistent for semantically equivalent inputs (see $\gP^x$ in Section~\ref{sec:task} and~\ref{sec:method}) as opposed to \textit{just} learning a new specific and isolated datapoint.
This consistency is indicative of an effective editing mechanism that taps into the knowledge stored in the model. 
However, not all datasets come with paraphrases of its inputs (\eg, in our case FEVER does not come with paraphrases and zsRE only has paraphrases for $30\%$ for the dataset). To this end, we generate semantically equivalent inputs using round-trip translation~\citep{sennrich-etal-2016-improving,wieting-gimpel-2018-paranmt}. 
We employ English-to-German and German-to-English Transformer models from Marian Neural Machine Translation~\citep[MarianNMT;][]{mariannmt} provided by Huggingface Transformers~\citep{wolf-etal-2020-transformers}. We use beam search with beam size $5$ to obtain $25$ paraphrases. 
From this set, we exclude any candidate paraphrase $\hat x$ of $x$ for which the prediction $\hat y$ supported by $f(\hat x; \theta)$ does not match the prediction $y$ supported by $f(x; \theta)$.
This filtering ensures that, according to the current model, all paraphrases have the exact same prediction.

\subsection{Architecture details}
The original models we want to modify are a BERT base model~\citep{devlin-etal-2019-bert} and a BART base model~\citep{lewis2019bart} for fact-checking and question answering respectively. They are both Transformer based models with 12 layers each and hidden size of 768. BERT has 12 heads, where BART has 16. They have 110M and 139M parameters respectively. BERT has a vocabulary size of 30,522 where BART has 50,265.

\editor has a small single-layered bidirectional-LSTM with input size 768 and hidden size of 128. The FFNN that condenses the LSTM states follows a [256, $\tanh$, 1024] architecture where the 5 FFNN have all a [1024, $\tanh$, $d$] architecture where $d$ depends on the weight to modify. In our experiments, we do not use our model to modify biases, layer norms, word and positional embeddings of LMs. Overall, \editor has 54M and 67M parameters for BERT and BART respectively.

\subsection{Training details}
The original models which we want to modify are trained with a batch size of 256 using Adam~\citep{kingma2014adam} (learning rate of 3e-5) with weight decay (1e-2) and a linear schedule with warm-up (50k total number of updates and 500 warm-up updates). We trained for a maximum  of 20 epochs and employ model selection using accuracy on the validation set.\footnote{We trained on 4 Nvidia Titian X 12GB which take approximately 10 minutes for FC and 3 hours for QA.}

\editor models are trained with a batch size of 1024 for FC and 256 for QA using Adam (learning rate of 3e-4 for the parameters and 1e-1 for the Lagrangian multiplier) with weight decay (1e-2) and a linear schedule with a warm-up (200k total number of updates and 1k warm-up updates). We trained for a maximum of 200 epochs and employ model selection using overall accuracy (success rate and retain accuracy) on the validation set (approximated using mini-batches).\footnote{We trained on 4 Nvidia Titian X 12GB which take approximately 1 day for FC and 3 days for QA.} The margin for the $\gC_{KL}$ is annealed between 1e-1 and 1e-3 for the fact-checking model, and between 1e-3 and 1e-5 for the BART question answering model. For the sequence-to-sequence loss, we employ a cross-entropy loss with label smoothing of 0.1.

\section{Additional Results} \label{app:additional}

\paragraph{Update Analysis}
During preliminary experiments, we studied a version of our hyper-network that did not exploit gradient information (see Equation~\ref{eq:hypernet}). Without gradient information, on FC the models converged $\approx\!10$ times slower to reach the same accuracy and did not converge for QA (\ie, the model was not able to get $>75\%$ success rate and $>50\%$ retain accuracy). That suggest that the gradients are helpful and actually used by our hyper-network but should not used directly, without a modification. To better show this, in Table~\ref{tab:cosine} we report correlations between different update methods and the gradient in terms of cosine similarities between updates. Naturally, fine-tuning and the gradient are highly correlated, but our method (with and without additional paraphrases supervision), poorly correlates with the others. Low cosine similarity can be due to two factors i) the model indeed projects the gradient to a different and more `knowledge preserving' direction, or ii) the parameter space is so large that cosine similarity gets to zero very quickly, not revealing the genuine underlying similarity. 

\begin{table}[t]
    \small
    \centering
    \begin{tabular}{l|cccc}
    \toprule
    {} & $\nabla_\theta \gL$ & Fine-tune & $\gC_{KL}$ & $\gC_{KL}$ + $\gP^x$ \\
    \midrule
    $\nabla_\theta \gL$     &  \phantom{-}1.000 &  \phantom{-}0.451 & -0.018 & -0.025 \\
    Fine-tune               &  \phantom{-}0.451 &  \phantom{-}1.000 & -0.010 & -0.011 \\
    $\gC_{KL}$              & -0.017 & -0.010 &  \phantom{-}1.000 &  \phantom{-}0.183 \\
    $\gC_{KL}$ + $\gP^x$    & -0.021 & -0.011 &  \phantom{-}0.183 &  \phantom{-}1.000 \\
    \bottomrule
\end{tabular}
    \caption{Average cosine similarities between different update methods and the gradient for the update as well. Fine-tuning is applied to all layers.}
    \label{tab:cosine}
\end{table}

\begin{figure}[t]
    \centering
    \begin{subfigure}[b]{0.48\textwidth}
        \centering
        \includegraphics[width=\textwidth]{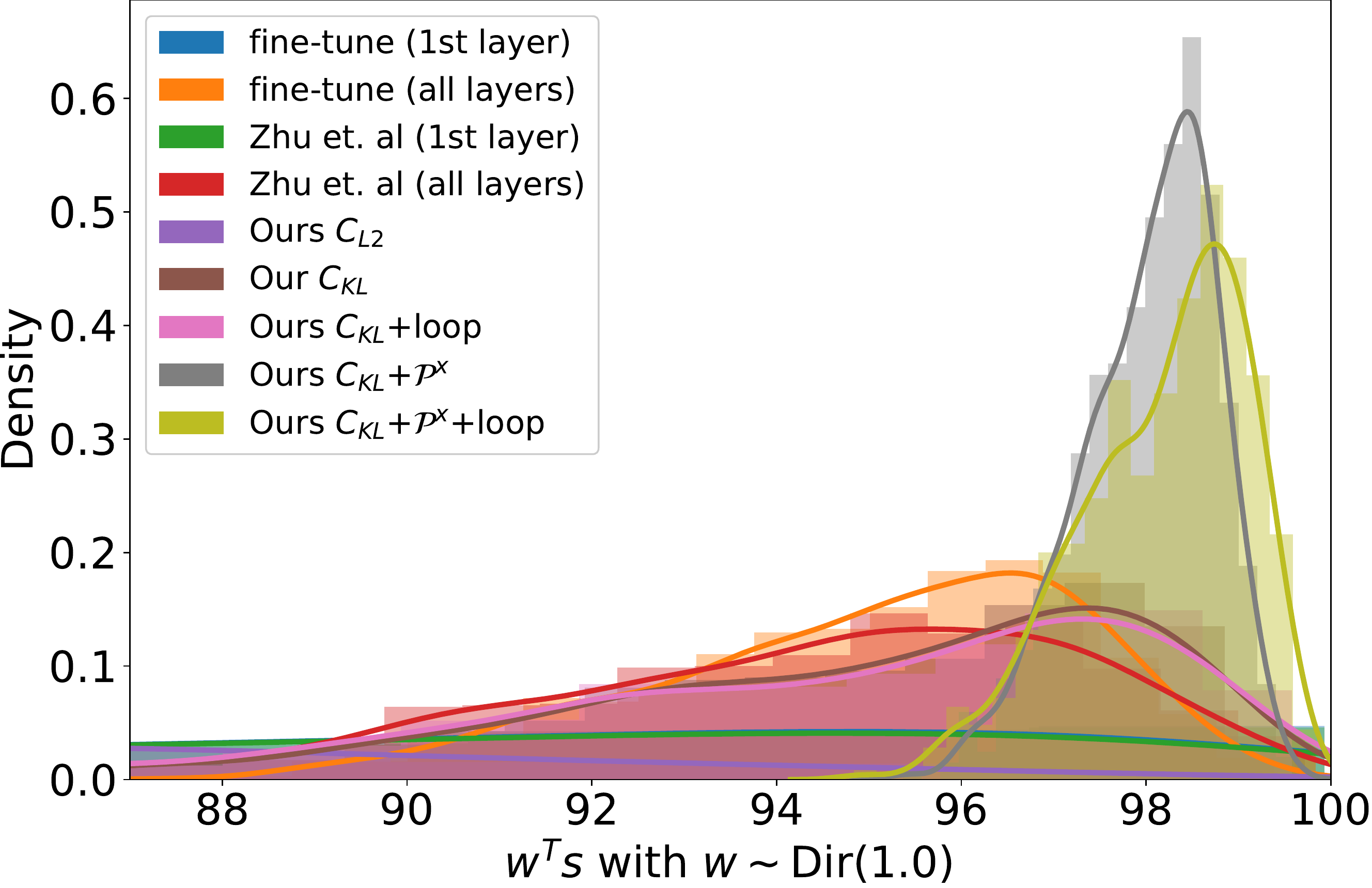}
        \caption{Fact-checking.}
        \label{fig:dirichlet-fc}
    \end{subfigure}
    \par\vspace{.5em}
    \begin{subfigure}[b]{0.48\textwidth}
        \centering
        \includegraphics[width=\textwidth]{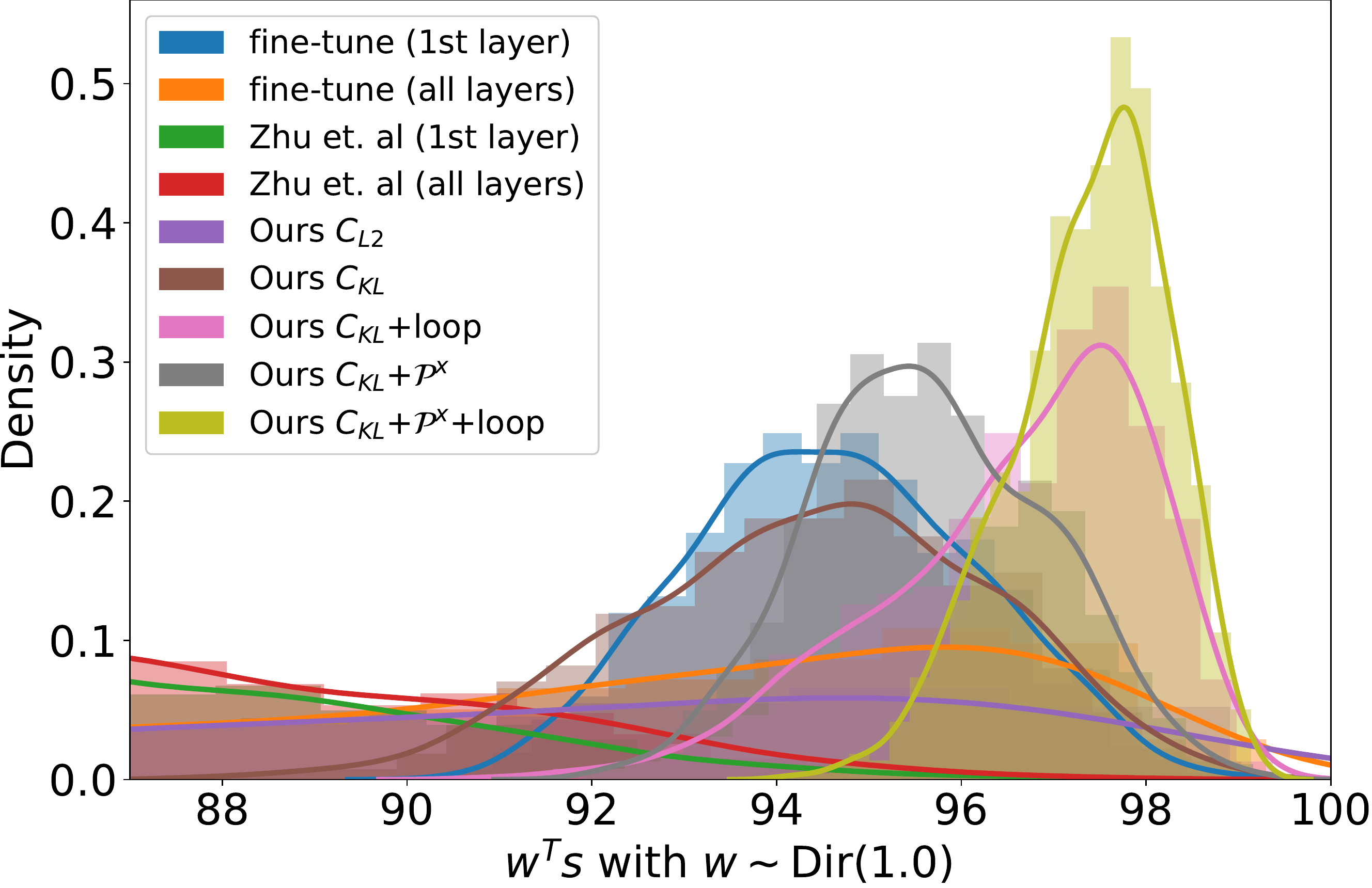}
        \caption{Question answering.}
        \label{fig:dirichlet-qa}
    \end{subfigure}
    \caption{Probability distributions of weighted sum of metrics according to 1k random assignments sampled from a Dirichlet distribution (with $\alpha=1$---see all values in Table~\ref{tab:fever-zsre}). Sampling weights allows to interpret the score in a probabilistic way. \editor (with different variants) presents distributions that are more skewed towards a high score (100) indicating that it is highly likely that when assigning some weights to the metrics, the weighted sum will be in favour of our method. \textit{Better view with colors.}}
    \label{fig:dirichlet}
\end{figure}

\begin{figure}[t]
    \centering
    \begin{subfigure}[b]{0.46\textwidth}
        \centering
        \includegraphics[width=\textwidth]{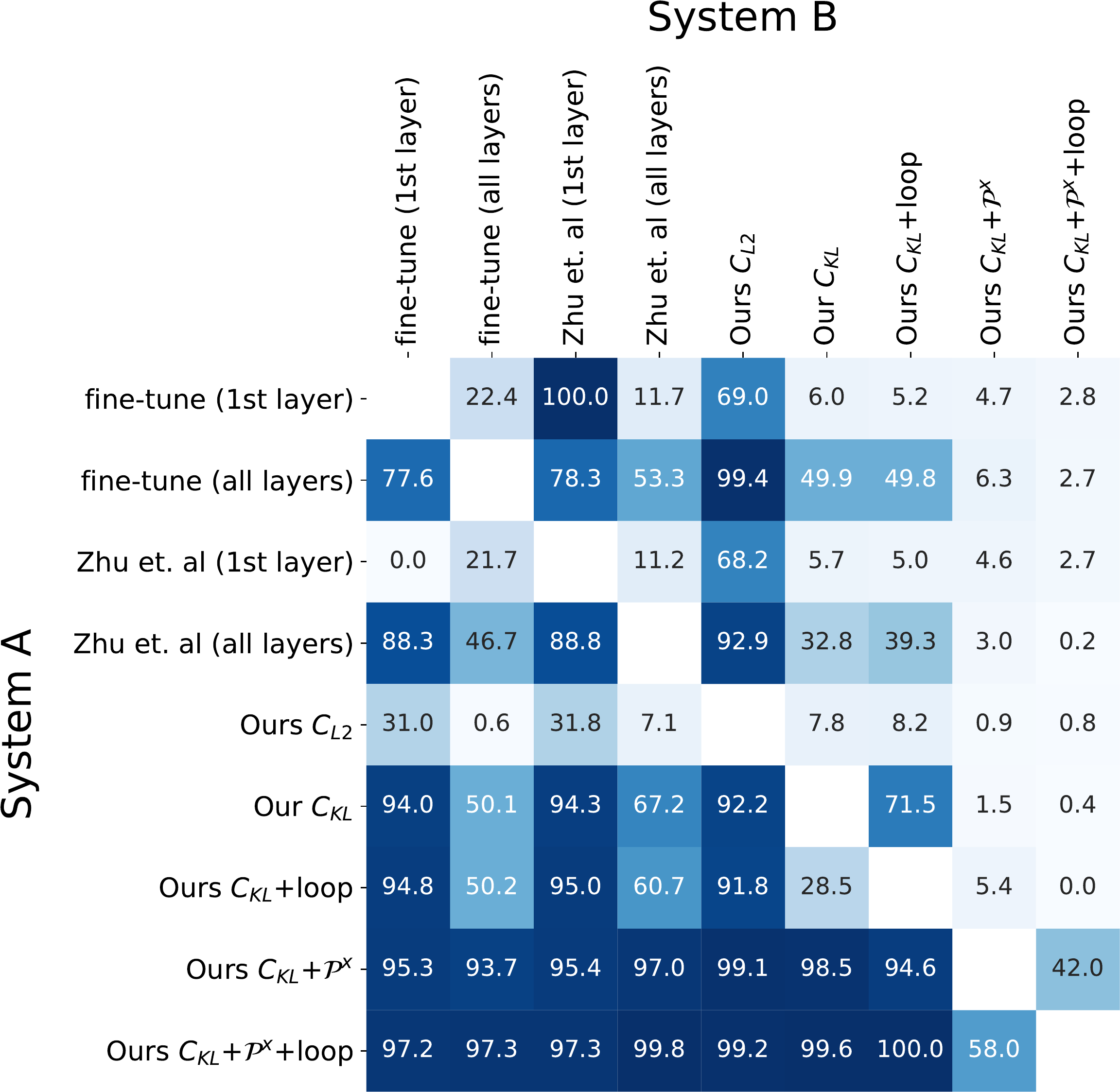}
        \caption{Fact-checking.}
        \label{fig:heatmap-fc}
    \end{subfigure}
    \par\vspace{.5em}
    \begin{subfigure}[b]{0.46\textwidth}
        \centering
        \includegraphics[width=\textwidth]{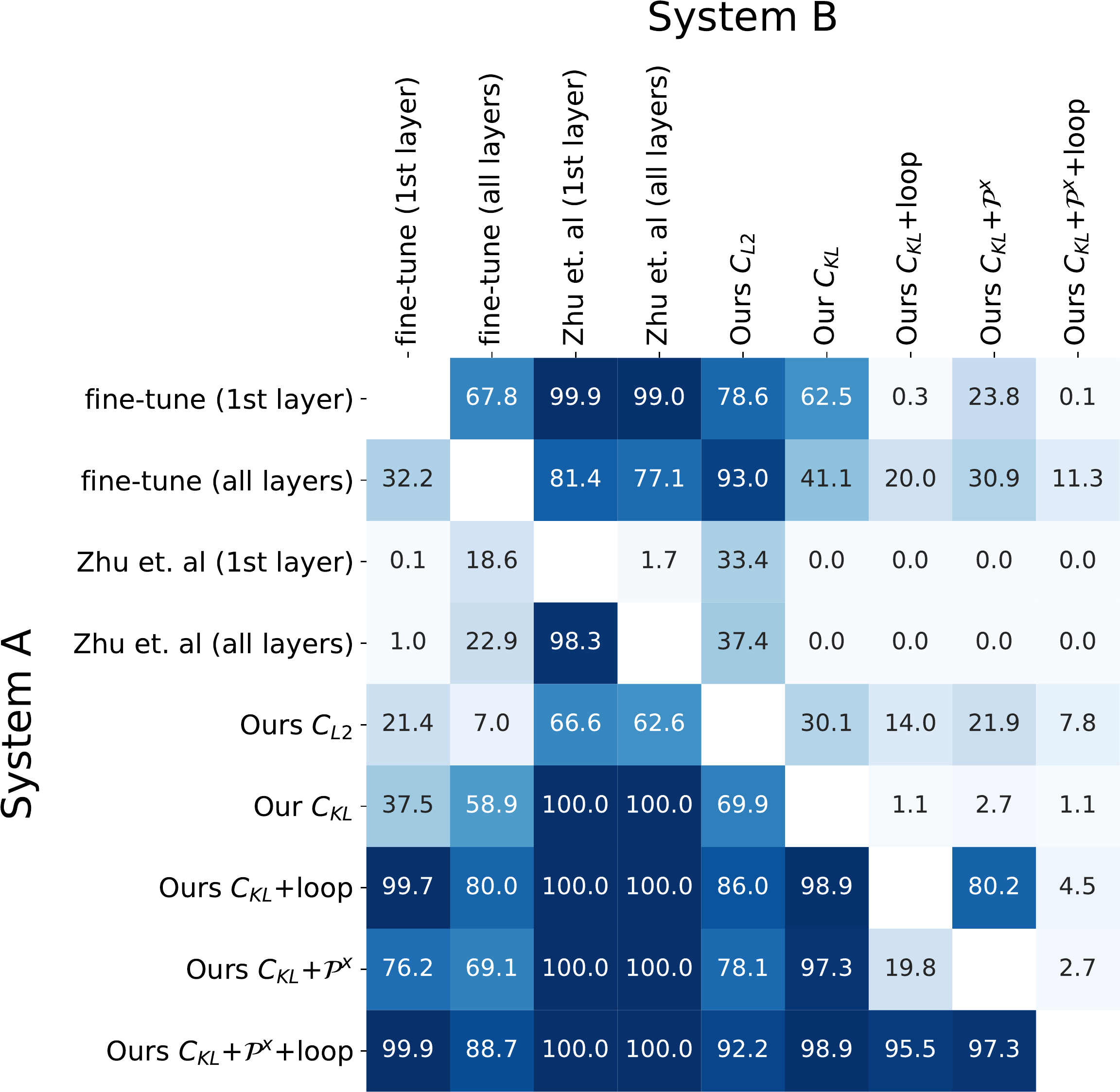}
        \caption{Question answering.}
        \label{fig:heatmap-qa}
    \end{subfigure}
    \caption{Probability that \textit{system A} is better than \textit{system B} according to a weighted sum of metrics (see individual values in Table~\ref{tab:fever-zsre}) sampling mixing coefficients $1,000$ times from a Dirichlet distribution (with $\alpha=1$ to cover a diverse spectrum of metric combinations). %
    The probability that \editor (with $\gC_{KL}$ + $\gP^x$ + loop) is better than competing systems is high ($>97\%$ for FC and $>88\%$ for QA) indicating that it is highly likely that when assigning some weights to the metrics, the weighted sum will be in favour of our method. \textit{Better view with colors.}}
    \label{fig:heatmap}
\end{figure}